\documentclass[10pt,twocolumn,letterpaper]{article}

\usepackage[arxiv]{iccv}      % To produce the REVIEW version
\usepackage{times}
\usepackage{graphicx}
\usepackage{amsmath, amssymb}
\usepackage{indentfirst}
\usepackage{multirow}
\usepackage{booktabs} % 美化表格线条
\usepackage{threeparttable} % 表格脚注支持
\usepackage{wrapfig}
\usepackage{xcolor}
\definecolor{iccvblue}{rgb}{0.21,0.49,0.74}
\usepackage[pagebackref,breaklinks,colorlinks,allcolors=iccvblue]{hyperref}
\def\paperID{1323} % *** Enter the Paper ID here
\def\confName{ICCV}
\def\confYear{2025}

\title{Learning Pixel-adaptive Multi-layer Perceptrons for Real-time Image Enhancement}

\author{Junyu Lou \qquad 
Xiaorui Zhao \qquad 
Kexuan Shi \qquad 
Shuhang Gu\thanks{Corresponding author}\\
University of Electronic Science and Technology of China\\
{\tt\small \{junyulou.jy, shuhanggu\}@gmail.com}\\
{\tt\small \href{https://github.com/LabShuHangGU/BPAM}{https://github.com/LabShuHangGU/BPAM}}
}

\begin{document}
\maketitle
\begin{abstract}

 Deep learning-based bilateral grid processing has emerged as a promising solution for image enhancement, inherently encoding spatial and intensity information while enabling efficient full-resolution processing through slicing operations. However, existing approaches are limited to linear affine transformations, hindering their ability to model complex color relationships. Meanwhile, while multi-layer perceptrons (MLPs) excel at non-linear mappings, traditional MLP-based methods employ globally shared parameters, which is hard to deal with localized variations. To overcome these dual challenges, we propose a \textbf{B}ilateral Grid-based \textbf{P}ixel-\textbf{A}daptive \textbf{M}ulti-layer Perceptron (BPAM) framework. Our approach synergizes the spatial modeling of bilateral grids with the non-linear capabilities of MLPs. Specifically, we generate bilateral grids containing MLP parameters, where each pixel dynamically retrieves its unique transformation parameters and obtain a distinct MLP for color mapping based on spatial coordinates and intensity values. In addition, we propose a novel grid decomposition strategy that categorizes MLP parameters into distinct types stored in separate subgrids. Multi-channel guidance maps are used to extract category-specific parameters from corresponding subgrids, ensuring effective utilization of color information during slicing while guiding precise parameter generation. Extensive experiments on public datasets demonstrate that our method outperforms state-of-the-art methods in performance while maintaining real-time processing capabilities.
\end{abstract}    
\section{Introduction}
\label{sec:intro}

Image enhancement is a fundamental task in computational photography and computer vision, aimed at improving the visual quality of images by adjusting attributes such as brightness, contrast, sharpness and exposure. With the widespread use of high-resolution displays and advanced imaging systems, there is an increasing demand for efficient techniques that can provide high-quality enhancements across various capture scenarios.
\begin{figure}[t]
  \flushright
  \includegraphics[width=1\linewidth]{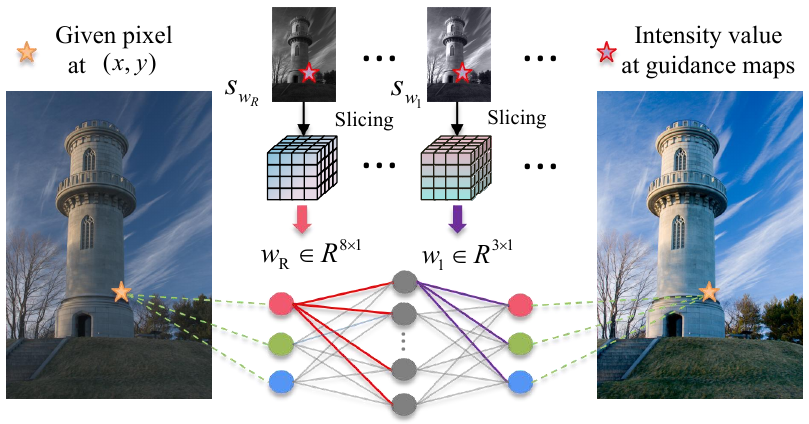}
  \caption{Pixel-adaptive MLP. $S$ represents the intensity value located at $(x, y)$ in corresponding guidance map. The red lines in the MLP represent the contributions of the input layer's R channel to the hidden layer, while the purple lines indicate the contributions of the first neuron in the hidden layer to the output layer. Each pixel on the image can obtain its unique MLP parameters based on its coordinates and pixel value.}
  \label{fig:figure1}
\end{figure}

Traditional image enhancement methods \cite{hwang2012context,lee2016automatic,kang2010personalization,liu2014autostyle,reinhard2001color} often rely on carefully designed algorithms that apply fixed transformations to pixel values. While effective to some extent, these approaches lack the flexibility to adapt to varying image content and complex real-world conditions. Recently, deep learning-based approaches \cite{chen2018learning,chen2018deep,ignatov2017dslr,jiang2021enlightengan,kim2020jsi,moran2020deeplpf,wei2018deep,wu2022uretinex,yan2016automatic,zhang2019kindling} have emerged, leveraging the powerful feature extraction and modeling capabilities of neural networks to facilitate sophisticated color transformations. Despite their success in performance, these methods tend to incur heavy computational costs, making real-time processing challenging and limiting their applicability in resource-constrained environments.

This challenge has spurred the emergence of hybrid approaches that synergize deep learning with computationally efficient physical models. Among these, 3D Lookup Table-based methods \cite{liu20234d,wang2021real,yang2022adaint,yang2022seplut,zeng2020learning,zhang2022clut,liu2024pixel,kim2024image,zhang2023lookup} have demonstrated particular success in image enhancement tasks. 3D LUTs offer a compact and fast means of mapping input color values to enhanced outputs through interpolation within a precomputed 3D lattice, and only need a lightweight network to predict weights for combining pre-defined LUTs, making them suitable for real-time applications. However, this computational advantage comes with inherent limitations: As image-level operators, conventional 3D LUT methods process colors in isolation from their spatial context, which may tend to produce spatially invariant enhancements. Although some spatial-aware 3D LUT approaches \cite{wang2021real,kim2024image} enhance preliminary processing with contextual features, they ultimately retain RGB-exclusive transformations during the critical color mapping phase.

Deep learning-based bilateral grid processing \cite{chen2016bilateral,gharbi2017deep,hashimoto2021fpga,wang2019underexposed,wang2023lgabl,xu2021bilateral,xu2021deep,zheng2021ultra,zhou20244k,chen2007real} represents another hybrid approach in computational imaging. Bilateral grid is a data structure that inherently incorporates spatial information while simultaneously capturing intensity-domain characteristics. HDRNet \cite{gharbi2017deep} pioneers the application of this structure in image enhancement pipelines, achieving a balance between spatial awareness and computational efficiency. The method utilizes a neural network to predict a bilateral grid containing affine coefficients in low resolution images. These coefficients are then mapped to the original image via slicing operation with learned guidance maps for pixel-wise affine transformations, enabling efficient feature propagation from low-resolution representations to full-resolution images. Although this framework unites spatial and intensity modeling, two critical limitations hinder its competitiveness against state-of-the-art 3D LUT methods. First, the reliance on affine transformations restricts the model to linear mappings, which may not adequate for capturing the complex, non-linear relationships required in diverse image enhancement scenarios. Second, inheriting the grayscale-oriented design of classical bilateral grids, HDRNet fuses input image to single-channel guidance map to extract all the coefficients. As a result, this strategy limits the model's ability to exploit multi-channel color information during the slicing operation.

To address the issues present in existing methods, we present a \textbf{B}ilateral Grid-based \textbf{P}ixel-\textbf{A}daptive \textbf{M}ulti-layer Perceptron (BPAM) framework, which leverages pixel-adaptive multi-layer perceptron in conjunction with bilateral grid to achieve high-quality, spatially aware color transformations with enhanced non-linear capabilities. Prior works \cite{he2020conditional} have proven that MLP is an effective function for modeling color transformation. However, the MLP used in current methods is only capable of image adaptation, applying the same parameters across the entire image. It often requires very deep hidden layers and large numbers of channels to ensure the generalization of model across diverse regions of the image. Despite this, they still struggle to address locally varying image properties.

Our model leverages the bilateral grid data structure, where a neural network generates bilateral grids containing MLP parameters from downsampled input image. Each pixel retrieves the corresponding MLP parameters from the grid based on its coordinates and pixel value, as shown in the right part of Figure \ref{fig:figure1}. In this approach, only a tiny MLP (3-8-3, denoting input-hidden-output layers) is required to achieve high performance. Furthermore, we observed that the parameters of MLP exhibit clear differences in their properties and can be categorized into different types. Accordingly, we partition the grid into multiple subgrids, with each one storing parameters corresponding to its specific category. When extracting coefficients, we use the information from each channel of guidance map to retrieve the associated coefficients from its corresponding subgrid, ensuring full utilization of the information across all channels and more accurate grid generation.
With non-linear and complex spatial-aware transformation, better utilization of color information, and lightweight network design, our approach could achieve high performance across multiple image enhancement tasks while meeting the requirements for real-time processing.

The main contributions of our paper are as follows:
\begin{itemize}
    \item We propose a pixel-adaptive MLP model based on the bilateral grid for modeling color transformations with high non-linear capability. Since the MLP parameters for each pixel are unique, our model dynamically adjusts color attributes on a per-pixel basis, capturing intricate local variations and improving overall image quality.
    \item We introduce a grid decomposition strategy where in the coefficients stored in the bilateral grid are decomposed into distinct components, this method seamlessly integrates with our MLP parameter estimation, ensuring the precise generation of information within the grid and the full utilization of each color channel.
    \item Extensive experiments on three publicly available datasets demonstrate the effectiveness and efficiency of the proposed method.
\end{itemize}

\section{Related Work}
\label{sec:related}

\subsection{Image Enhancement Methods}

Recent advances in image enhancement have diverged into two principal methodologies. The first paradigm \cite{chen2018learning,chen2018deep,ignatov2017dslr,jiang2021enlightengan,kim2020jsi,moran2020deeplpf,wei2018deep,wu2022uretinex,yan2016automatic,zhang2019kindling} employs end-to-end deep learning architectures, particularly fully convolutional networks, to learn direct pixel-to-pixel transformations. While delivering impressive enhancement quality, these models typically necessitate deep, complex structures with excessive parameterization, incurring substantial computational and memory overhead that hinders real-time deployment.

Meanwhile, a competing hybrid paradigm has emerged, synergizing deep learning with computationally efficient physical models. These methods leverage streamlined neural networks to predict parameters for established enhancement components—including mapping curve \cite{guo2020zero,kim2020global,li2020flexible,moran2021curl,park2018distort,song2021starenhancer}, MLP \cite{he2020conditional}, 3D LUT, and bilateral grid. 

3D LUT, in particular, has gained traction as an efficient alternative for color transformation tasks. It offers a competitive performance on enhancement tasks with fast processing speeds. Since Zeng \textit{et al.} first introduced learnable 3D LUT \cite{zeng2020learning}, numerous variants \cite{liu20234d,wang2021real,yang2022adaint,yang2022seplut,zhang2022clut,liu2024pixel,kim2024image,zhang2023lookup} have been proposed. Some methods attempt to enhance the spatial modeling capabilities of 3D LUT. SA-3DLUT~\cite{wang2021real} enhances adaptability through pixel-type classification and dual-head weight prediction for LUTs fusion, but dramatically increases parameter count. LUTwithBGrid~\cite{kim2024image} integrates bilateral grids to generate spatial information before mapping to 3D LUTs, but the bilateral grids used during slicing do not compute gradients of spatial coordinates. Essentially, it only serves as a 1D lookup table, limiting its effectiveness in handling spatial features.

Bilateral grid is a data structure that captures both spatial and intensity information. In some deep learning-based approaches \cite{gharbi2017deep,hashimoto2021fpga,wang2019underexposed,wang2023lgabl,xu2021bilateral,xu2021deep,zheng2021ultra,zhou20244k,chen2007real}, bilateral grids are constructed from input images and transformed into task-specific feature representations through a slicing operation to facilitate downstream processing. Most these methods follow the processing pipeline of \cite{gharbi2017deep}, where they apply affine transformations to obtain the required features and employ a single-channel guidance map to guide the slicing process. We believe that the full potential of bilateral grids in the field of image enhancement has yet to be explored.

\subsection{MLP-based Architectures for Visual Processing}

Multi-layer perceptrons offer strong nonlinear modeling capabilities and also maintain relatively simple architectures, making them widely used in computer vision tasks. Recent advances in MLP-based approaches have demonstrated remarkable progress across various vision domains. For instance, MLP-Mixer \cite{tolstikhin2021mlp} partitions images into patches and applies separate MLPs for spatial and channel mixing, while ResMLP \cite{touvron2022resmlp} introduces residual connections and GELU activations to enhance stability. gMLP \cite{liu2021pay} incorporates gating mechanisms to improve spatial modeling, and MAXIM \cite{tu2022maxim} adopts a multi-axis design to handle a range of image processing tasks. CycleMLP \cite{chen2023cyclemlp} leverages cyclic shifts to fuse local and global features, whereas Hire-MLP \cite{guo2022hire} employs a hierarchical structure to preserve fine details in high-resolution images. Meanwhile, Vision Permutator \cite{hou2022vision} applies permutation operations to pre-mix spatial information before MLP processing.

Beyond architectural innovations, Ding \textit{et al.} \cite{ding2021repmlp} proposed a re-parameterization technique to boost the capability of MLPs, and Shi \textit{et al.} \cite{shi2024improved} introduced a Fourier-based re-parameterization method to improve the spectral bias of MLPs. In the field of image enhancement, CSRNet \cite{he2020conditional} replaces traditional color transformation into MLPs for efficient global image retouching.

In this work, the MLPs parameters are predicted using the calculated bilateral grids, which ultimately enables spatially aware image enhancement.
\section{Method}
\label{sec:method}
\begin{figure*}[t]
  \centering
  \includegraphics[width=\textwidth]{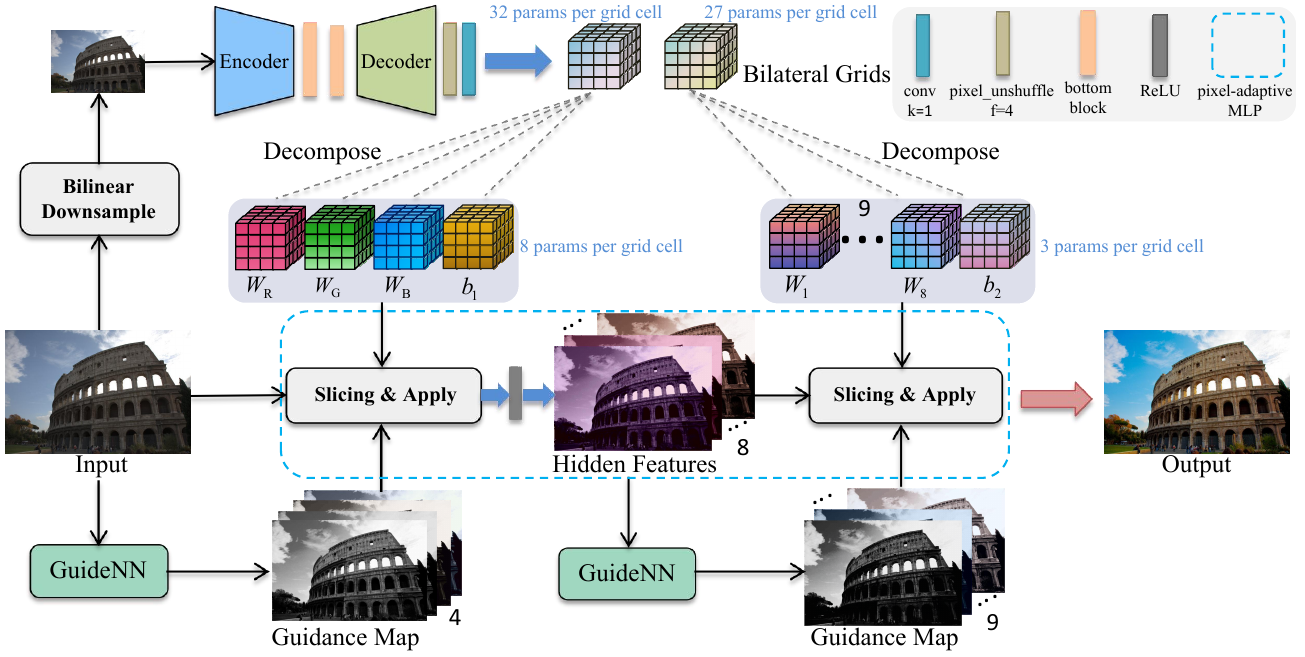}
  \caption{Illustration of proposed bilateral grid-based pixel-adaptive multi-layer perceptron (BPAM) framework. It comprises three main components, 1) the backbone for generating the grids, 2) the GuidNN network for generating the guidance maps, and 3) the MLP for the image transformation. Regarding the third component, we combine the generated guidance maps with the grids to perform two slicing operations, thereby obtaining the parameters for both parts of the MLP to transform the image. For more details, please refer to Figure \ref{fig:figure3}.}
  \label{fig:figure2}
  \vspace{-1em}
\end{figure*}
The pipeline for BPAM is shown in Figure \ref{fig:figure2}. We first downsample the image and utilize our backbone to extract features from the low-resolution version. Then, we employ pixel unshuffle to further reduce the grid size while increasing the number of channels, followed by a 1×1 convolution to obtain the final grids. Additionally, since the MLP parameters consist of two parts, we sequentially generate two sets of guidance maps and perform two slicing operations to form a complete MLP for transforming the image.

\subsection{Preliminary: Bilateral Grid}

The bilateral grid is a 3D volumetric structure designed to jointly encode spatial coordinates and pixel intensity, enabling efficient edge-aware image processing. Because bilateral grid strikes a good balance between efficiency and effectiveness, several methods have adopted this data structure in their models for their purpose.

Given an input image \(I \in \mathbb{R}^{W\times H\times 3}\), they employ convolutional neural networks to extract features on image and generate a bilateral grid $\mathcal{G}\in\mathbb{R}^{W/s_x\times H/s_y\times L/s_r}$, where $L$ represents the range of pixel values (typically 255), $s_x$, $s_y$ and $s_r$ control spatial and intensity downsampling strides. Each grid cell $\mathcal{G}(i,j,k)$ aggregates features from pixels within a spatial window and an intensity range.
This coarsely discretized representation reduces computational complexity while preserving structural edges through intensity-domain decoupling.

Once the bilateral grid is constructed, the slicing operation \cite{chen2007real} is used to apply the parameters or coefficients contained in the grid to the original image and reconstruct a full-resolution output. Previous methods \cite{chen2016bilateral,gharbi2017deep} generate a single-channel guidance map to serve as the intensity value and direct the slicing operation. The guidance map $g$ of size $H\times W$ is  first processed by lifting each pixel at position \((x,y)\) with intensity \(I(x,y)\) into a three-dimensional space using the mapping:
\begin{equation}
(u,v,r) = \left(\frac{x}{s_x},\,\frac{y}{s_y},\,\frac{I(x,y)}{s_r}\right).
\end{equation}

For a given pixel \(p\) at \((x,y)\) with corresponding continuous coordinate \((u,v,r)\), we express:
\begin{equation}
u = i_0 + \delta_u,\quad v = j_0 + \delta_v,\quad r = k_0 + \delta_r,
\end{equation}
where \(i_0 = \lfloor u \rfloor\), \(j_0 = \lfloor v \rfloor\), \(k_0 = \lfloor r \rfloor\) are the integer parts and \(\delta_u\), \(\delta_v\), \(\delta_r\) are the fractional parts. The output \(A(x,y)\) is then obtained via trilinear interpolation over the eight neighboring grid cells:
\begin{equation}
A(x,y) = \sum_{a=0}^{1} \sum_{b=0}^{1} \sum_{c=0}^{1} w_{abc}\, \mathcal{G}(i_0+a,\,j_0+b,\,k_0+c).
\end{equation}

Here, the weights are defined as:
\begin{equation}
w_{abc} = (1-\delta_u)^{\,1-a}\,\delta_u^{\,a}\,(1-\delta_v)^{\,1-b}\,\delta_v^{\,b}\,(1-\delta_r)^{\,1-c}\,\delta_r^{\,c}.
\end{equation}

In this compact expression, the interpolation weights for each neighboring grid cell are directly determined by the fractional offsets, ensuring that:
\begin{equation}
\sum_{a,b,c \in \{0,1\}} (1-\delta_u)^{1-a}\delta_u^a\,(1-\delta_v)^{1-b}\delta_v^b\,(1-\delta_r)^{1-c}\delta_r^c = 1.
\end{equation}

Since the slicing operation is fully differentiable, it permits end-to-end gradient back propagation through the grid, thereby enabling the entire enhancement network to be trained jointly.
Methods like HDRNet \cite{gharbi2017deep} employ the slicing operation to obtain the corresponding affine coefficients \(A(x,y)\)$\in\mathbb{R}^{3\times4}$ for each pixel in the original image. Each pixel’s output is computed as:
\begin{equation}
O(x,y)=\alpha(x,y,s)\cdot I(x,y)+\beta(x,y,s),
\end{equation}
where $s$ is the value at the coordinates $(x,y)$ on the guidance map, $I(x,y)\in\mathbb{R}^3$ is the input color vector, $O(x,y)\in\mathbb{R}^3$ is the output color vector, $\alpha(x,y,s)\in\mathbb{R}^{3\times3}$ is a linear transformation matrix, $\beta(x,y,s)\in\mathbb{R}^{3}$ is a bias vector.

\subsection{Pixel-adaptive MLP Learning}
Although affine transformations are computationally efficient, their expressive power is limited because they cannot model complex non-linear mappings. So, we choose to apply a distinct MLP for each pixel to model the color mapping, our generated grids store the MLP parameters required for different regions.

As a general and effective color mapping function, MLP is typically considered a global operation. We believe this is because the MLPs used in previous work apply the same parameters across the entire image area. If the MLP parameters can be spatially adaptive, the situation would be completely different.

In our framework, we employ a three-layer U-net-style backbone to generate two bilateral grids that contain MLP parameters. The U-Net-style architecture enlarges the receptive field, which is particularly beneficial for image enhancement tasks. For every pixel on the image, we form a unique three-layer MLP (3-8-3) to perform the transformation, with the MLP parameters derived from the generated grids. Using these spatially-varying parameters, the MLP transformation is computed in two stages:
\begin{equation}
\mathbf{z}(x,y) = \sigma\left(\mathbf{W}_1(x,y,s) \cdot \mathbf{I}(x,y) + \mathbf{b}_1(x,y,s)\right).
\end{equation}

For the first stage, $I(x,y)\in\mathbb{R}^3$ is the input color vector, $z(x,y)\in\mathbb{R}^8$ is the hidden layer vector, $\mathbf{W}_1(x,y,s)\in\mathbb{R}^{8\times3}$ is the weight matrix, $\mathbf{b}_1(x,y,s)\in\mathbb{R}^{8}$ is the bias vector, $\sigma(\cdot)$ is a nonlinear activation function ReLU.
\begin{equation}
O(x,y) = \mathbf{W}_2(x,y,s) \cdot \mathbf{z}(x,y) + \mathbf{b}_2(x,y,s).
\end{equation}

For the second stage, $O(x,y)\in\mathbb{R}^3$ is the output color vector, $\mathbf{W}_2(x,y,s)\in\mathbb{R}^{3\times8}$ is the weight matrix, $\mathbf{b}_2(x,y,s)\in\mathbb{R}^{3}$ is the bias vector.

Specifically, each grid cell in the first grid contains 32 parameters, corresponding to the 24 weights and 8 biases required for generating the hidden layer of the MLP. Similarly, each grid cell in the second grid contains 27 parameters, corresponding to the 24 weights and 3 biases required for the last layer of the MLP.

It should be noticed that the parameters in the MLP are divided into two parts with a sequential order. To handle this, we generated two separate sets of guidance maps to guide the slicing operation for each part. Following the setup in HDRNet \cite{gharbi2017deep}, we use two convolutional networks to generate these guidance maps. The first network takes the original image as input, and the output guidance maps directs the slicing operation in the first grid to obtain the parameters required for generating the hidden layers. Then, we feed the hidden layers vector into the second network to produce the second set of guidance maps, which directs the slicing operation in the second grid to obtain the parameters necessary for generating the final result.

Through these designs, each pixel on the image can obtain its unique, simple yet fully functional MLP for color transformation.

\subsection{Grid Decomposition Strategy}

The bilateral grid was initially designed for single-channel grayscale images. To apply it to RGB images, one approach is to extend the bilateral grid to 5 dimensions $(X,Y,R,G,B)$, but this results in a significant increase in computational complexity. Therefore, previous methods \cite{chen2016bilateral,gharbi2017deep} adopted a solution in which the color information from the three channels is fused into a single channel to serve as the intensity value within the bilateral grid. They use a single-channel map to extract all the coefficients, and we believe this approach may lead to insufficient utilization of color information during the slicing operation. In contrast, 3DLUT can leverage all the color information when performing color mapping.

To tackle this issue, we introduce a grid decomposition strategy. Recognizing that the MLP parameters naturally fall into distinct groups, we partition the grid into several subgrids, each dedicated to a different category. For example, the first grid is segmented into four subgrids that each contain the weight for a specific color channel along with a shared bias. Similarly, the second grid is divided into nine subgrids, with each one storing the weight for a corresponding hidden channel and its shared bias.

\begin{figure}[t]
  \flushright
  \includegraphics[width=1\linewidth]{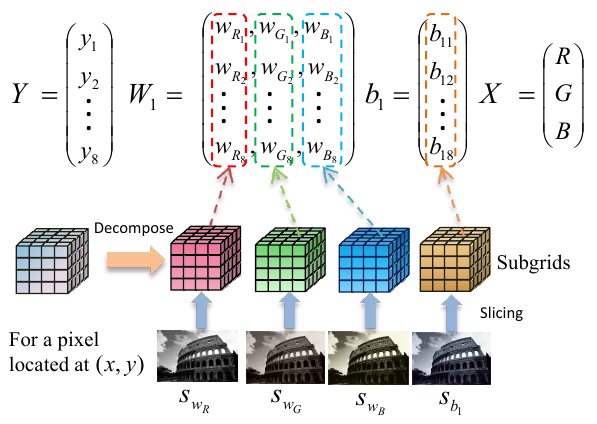}
  \caption{Illustration of grid decomposition for the first grid, which partition the MLP parameters of different categories into separate grids. Given the coordinates, we obtain the intensity value at the corresponding position in the guidance maps. $X$ is the input vector, $Y$ is the hidden layer verctor. The decomposition strategy for the second grid is similar. }
  \label{fig:figure3}
\end{figure}

This reorganization calls for specialized guidance map to steer the slicing process for each subgrid. Accordingly, we generate two multi-channel guidance maps: one comprising four channels and the other containing nine channels, with each channel aligning with its respective subgrid. Each guidance channel is then applied to extract parameters from its designated subgrid, as illustrated in Figure ~\ref{fig:figure3}.

\subsection{Loss Function}

Our loss function consists of three parts: MSE loss $\mathcal{L}_{2}$, SSIM loss $\mathcal{L}_{ssim}$, and perceptual loss $\mathcal{L}_{per}$.The perceptual loss $\mathcal{L}_{per}$ is the difference between ground truth and output image features on the pre-trained VGG19 \cite{simonyan2014very}. The final loss is empirically set as:
\begin{equation}
\mathcal{L}=\mathcal{L}_{2}+0.5\times\mathcal{L}_{ssim}+0.005\times\mathcal{L}_{per}.
\end{equation}
\section{Experiments}
\label{sec:experiments}

\begin{table*}[!t]
\centering
\fontsize{8pt}{10pt}\selectfont
\renewcommand{\arraystretch}{0.9}
\setlength{\tabcolsep}{8pt}
\resizebox{0.9\textwidth}{!}{%
\begin{tabular}{l|c|ccc|ccc}
\toprule
\multirow{2}{*}{\textbf{Methods}} & \multirow{2}{*}{\textbf{\#Params}} & \multicolumn{3}{c|}{\textbf{480p}} & \multicolumn{3}{c}{\textbf{Full Resolution}} \\
\cmidrule(lr){3-5}\cmidrule(lr){6-8}
 &  & \textbf{PSNR}$\uparrow$ & \textbf{SSIM}$\uparrow$ & $\boldsymbol{\triangle E}\downarrow$ & \textbf{PSNR}$\uparrow$ & \textbf{SSIM}$\uparrow$ & $\boldsymbol{\triangle E}\downarrow$ \\
\midrule
UPE \cite{wang2019underexposed}       & 927K   & 21.56 & 0.837 & 12.29 & 20.41 & 0.803 & 12.50 \\
DeepLPF \cite{moran2020deeplpf}   & 1.7M   & 24.97 & 0.897 & 6.22  & -     & -     & - \\ 
HDRNet \cite{gharbi2017deep}   & 482K   & 24.52 & 0.915 & 8.14  & 24.17 & 0.919 & 8.91 \\
CSRNet \cite{he2020conditional}   & 37K    & 25.19 & 0.921 & 7.63  & 24.23 & 0.920 & 8.75 \\
3DLUT \cite{zeng2020learning}    & 592K   & 25.07 & 0.920 & 7.55  & 24.39 & 0.923 & 8.33 \\
SA-3DLUT\textsuperscript{*} \cite{wang2021real} & 4.5M   & 24.67 & /     & /     & 24.27 & /     & / \\
LLF-LUT\textsuperscript{*} \cite{zhang2023lookup}  & 731K   & 25.53 & /     & /     & 24.52 & /     & / \\
LutBGrid \cite{kim2024image} & 464K & \textcolor{blue}{25.59} & \textcolor{blue}{0.932} & \textcolor{blue}{7.14} & \textcolor{blue}{24.57} & \textcolor{blue}{0.931} & \textcolor{blue}{8.03} \\
\midrule
\textbf{Ours} & 624K & \textcolor{red}{25.83} & \textcolor{red}{0.941} & \textcolor{red}{7.00} & \textcolor{red}{25.12} & \textcolor{red}{0.934} & \textcolor{red}{7.73} \\
\bottomrule
\end{tabular}
}
\caption{Quantitative comparison on the \textbf{FiveK} dataset for \textbf{tone mapping}. “-" means the result is not available due to insufficient GPU memory. “*” indicates that the results are obtained from their paper and “/” are absent results due to the unavailable code.}
\vspace{-3mm}
\label{tab:table1}
\end{table*}

\begin{table}[htbp]
\centering
\scriptsize
\label{tab:fivek-comparison}
\renewcommand{\arraystretch}{0.9}
\setlength{\tabcolsep}{8pt}
\resizebox{0.45\textwidth}{!}{
\begin{tabular}{l|ccc}
\toprule
\textbf{Methods} & \multicolumn{3}{c}{\textbf{480p}} \\
\cmidrule(lr){2-4}
                 & \textbf{PSNR}$\uparrow$ & \textbf{SSIM}$\uparrow$ & $\boldsymbol{\triangle E}\downarrow$ \\
\midrule
UPE \cite{wang2019underexposed}       & 21.88 & 0.853 & 10.80 \\
HDRNet \cite{gharbi2017deep}   & 24.66 & 0.915 & 8.06  \\
CSRNet \cite{he2020conditional}   & 25.17 & 0.924 & 7.75  \\
DeepLPF \cite{moran2020deeplpf}  & 24.73 & 0.916 & 7.99  \\
3DLUT \cite{zeng2020learning}    & 25.29 & 0.923 & 7.55  \\
SA-3DLUT \cite{wang2021real} & 25.50 & /     & /     \\
SepLUT\textsuperscript{*} \cite{yang2022seplut}   & 25.47 & 0.921 & 7.54  \\
AdaInt \cite{yang2022adaint}   & 25.49 & 0.926 & 7.47  \\
PLSA\textsuperscript{*} \cite{liu2024pixel}    & 25.57 & \textcolor{blue}{0.931} & 7.32 \\
LutBGrid \cite{kim2024image} & \textcolor{blue}{25.66} & 0.930 & \textcolor{blue}{7.29} \\
\midrule
\textbf{Ours} & \textcolor{red}{25.73} & \textcolor{red}{0.937} & \textcolor{red}{7.19} \\
\bottomrule
\end{tabular}
}
\caption{Quantitative comparison on the \textbf{FiveK} dataset (480p) for \textbf{photo retouching}.}
\vspace{-3mm}
\label{tab:table2}
\end{table}

\subsection{Datasets}

We evaluate our method on three datasets: MIT-Adobe FiveK \cite{bychkovsky2011learning}, PPR10K \cite{liang2021ppr10k} and LCDP \cite{wang2022local}.
The FiveK dataset is one of the most widely used benchmarks for enhancement tasks. It comprises 5,000 images captured with a variety of cameras. Each image has been manually retouched by five different expert photographers, we follow the practice in recent works \cite{zeng2020learning,yang2022adaint,kim2024image} to adopt the version retouched by expert C as the ground truth. We divide the dataset into 4,500 image pairs for training and 500 image pairs for testing, following the splitting approach used by \cite{zeng2020learning}.
The PPR10K dataset is designed specifically for portrait photo retouching. It contains 11,161 portrait images along with three professionally retouched counterparts. We perform three sets of experiments for each version of ground truth (A/B/C). Following the official split \cite{liang2021ppr10k}, we split the dataset into 8,875 image pairs for training and 2,286 image pairs for testing. The LCDP is a non-uniform illumination dataset, which displays non-uniform illumination caused by both 
\begin{table}[htbp]
\centering
\resizebox{0.45\textwidth}{!}{%
\begin{tabular}{c|c|cccc}
\toprule
\textbf{Dataset} & \textbf{Methods} & \textbf{PSNR}$\uparrow$ & $\boldsymbol{\triangle E}\downarrow$ & \textbf{PSNR}$^{H\!C}$$\uparrow$ & $\Delta E^{H\!C}\downarrow$ \\
\midrule

\multirow{6}{*}{\begin{tabular}[c]{@{}c@{}}PPR10K-a\end{tabular}}
  & 3DLUT \cite{zeng2020learning}            & 25.99 & 6.76 & 28.29 & 4.38 \\
  & SepLUT\textsuperscript{*} \cite{yang2022seplut}            & 26.28 & 6.59 & /     & /    \\
  & AdaInt \cite{yang2022adaint}            & 26.33 & 6.56 & 29.57 & 4.26 \\
  & PLSA\textsuperscript{*} \cite{liu2024pixel}              & \textcolor{blue}{25.50} & 6.58 & \textcolor{blue}{29.87} & / \\
  & LutBGrid \cite{kim2024image}          & 26.45 & \textcolor{blue}{6.51} & 29.70 & \textcolor{blue}{4.23} \\
  & Ours              & \textcolor{red}{26.58} & \textcolor{red}{6.40} & \textcolor{red}{29.82} & \textcolor{red}{4.17} \\
\midrule

\multirow{6}{*}{\begin{tabular}[c]{@{}c@{}}PPR10K-b\end{tabular}}
  & 3DLUT \cite{zeng2020learning}             & 25.06 & 7.51 & 28.36 & 4.85 \\
  & SepLUT\textsuperscript{*} \cite{yang2022seplut}            & 25.23 & 7.49 & /     & /    \\
  & AdaInt \cite{yang2022adaint}            & 25.40 & 7.33 & 28.65 & 4.75 \\
  & PLSA\textsuperscript{*} \cite{liu2024pixel}              & \textcolor{blue}{25.51} & 7.29 & \textcolor{blue}{28.80} & / \\
  & LutBGrid \cite{kim2024image}          & 25.48 & \textcolor{blue}{7.19} & 28.72 & \textcolor{blue}{4.66} \\
  & Ours              & \textcolor{red}{25.57} & \textcolor{red}{7.18} & \textcolor{red}{28.84} & \textcolor{red}{4.66} \\
\midrule

\multirow{6}{*}{\begin{tabular}[c]{@{}c@{}}PPR10K-c\end{tabular}}
  & 3DLUT \cite{zeng2020learning}             & 25.46 & 7.43 & 28.80 & 4.82 \\
  & SepLUT\textsuperscript{*} \cite{yang2022seplut}            & 25.59 & 7.51 & /     & /    \\
  & AdaInt \cite{yang2022adaint}            & 25.68 & 7.31 & 28.93 & 4.76 \\
  & PLSA\textsuperscript{*} \cite{liu2024pixel}              & \textcolor{blue}{25.76} & \textcolor{blue}{7.27} & \textcolor{blue}{29.01} & / \\
  & LutBGrid \cite{kim2024image}          & 25.72 & 7.28 & 29.00 & \textcolor{blue}{4.72} \\
  & Ours              & \textcolor{red}{25.85} & \textcolor{red}{7.22} & \textcolor{red}{29.09} & \textcolor{red}{4.69} \\
\bottomrule
\end{tabular}
}
\caption{Quantitative comparison on the \textbf{PPR10K} dataset for \textbf{photo retouching}.}
\label{tab:table3}
\end{table}
overexposure and underexposure within individual images. It consists of 1,415 training images, 100 validation images, and 218 test images.

We perform three common tasks in the field of image enhancement: tone mapping, photo retouching, and exposure correction. For tone mapping, we utilize the FiveK dataset at both 480p and original resolutions, processing 16-bit CIE XYZ inputs to 8-bit sRGB outputs. The photo retouching experiments involve two configurations: FiveK at 480p resolution (8-bit sRGB to 8-bit sRGB) and PPR10K at 360p resolution (16-bit sRGB to 8-bit sRGB). Exposure correction task is performed on the LCDP dataset at original resolution (8-bit sRGB to 8-bit sRGB).

\subsection{Implementation Details}
\begin{figure*}[t]
  \centering
  \setlength{\belowcaptionskip}{0pt}
  \includegraphics[width=\textwidth]{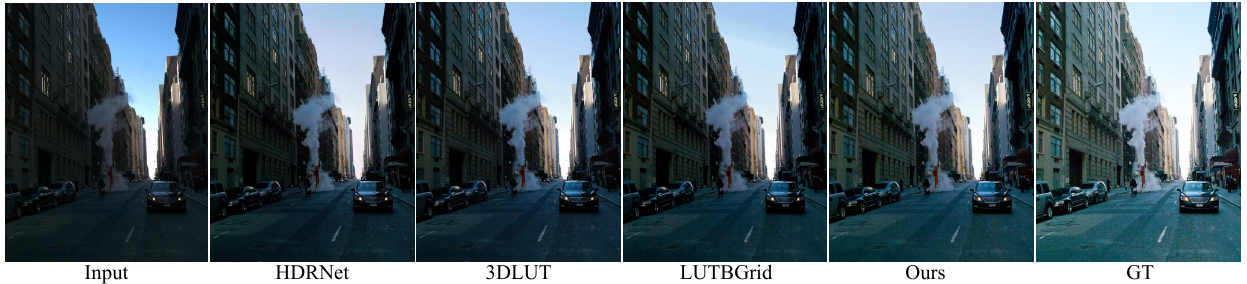}
  \caption{Visual comparison with state-of-the-art methods on the \textbf{FiveK} dataset (480p) for \textbf{photo retouching}.}
  \label{fig:figure4}
\end{figure*}

\begin{figure*}[t]
  \centering
  \setlength{\belowcaptionskip}{0pt}
  \includegraphics[width=\textwidth]{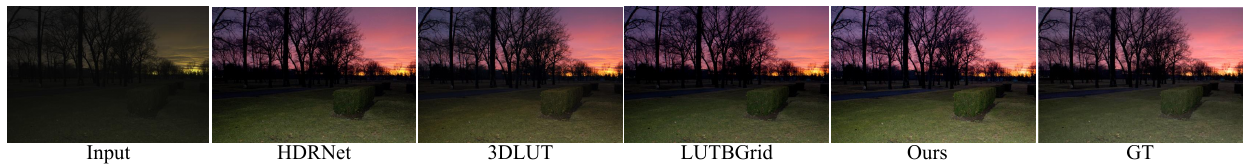}
  \caption{Visual comparison with state-of-the-art methods on the\textbf{ FiveK} dataset (480p) for \textbf{tone mapping}.}
  \label{fig:figure5}
\end{figure*}

\begin{table}[htbp]
\centering
\renewcommand{\arraystretch}{1.2} % 行高
\setlength{\tabcolsep}{8pt}       % 列间距
\resizebox{0.45\textwidth}{!}{%
\begin{tabular}{l|c|ccc}
\toprule
\textbf{Methods}   & \#\textbf{Params} & \textbf{PSNR$\uparrow$} & \textbf{SSIM$\uparrow$} & \textbf{LPIPS$\downarrow$} \\
\midrule
RetinexNet \cite{wei2018deep}   & 840K  & 16.20 & 0.630 & 0.294 \\
URetinexNet \cite{wu2022uretinex}  & 1.32M & 17.67 & 0.737 & 0.250 \\
DRBN \cite{xie2023semi}        & 580K  & 15.47 & 0.698 & 0.315 \\
SID  \cite{chen2018learning}        & 7.4M  & 21.89 & 0.808 & 0.178 \\
ZeroDCE \cite{guo2020zero}     & 79K   & 18.96 & 0.774 & 0.206 \\
ENC-SID \cite{huang2022exposure}     & 7.45M & 22.66 & 0.820 & 0.163 \\
CLIP-LIT \cite{liang2023iterative}    & 280K  & 19.24 & 0.748 & 0.226 \\
FECNet  \cite{huang2022deep}     & 150K  & 22.34 & 0.804 & 0.233 \\
LUTBGrid \cite{kim2024image}    & 464K  & 22.71 & 0.803 & 0.154 \\
ENC-DRBN \cite{huang2022exposure}     & 580K  & 23.08 & 0.830 & 0.154 \\
LCDPNet \cite{wang2022local}      & 960K  & 23.24 & 0.842 & 0.137 \\
CoTF \cite{li2024real}        & 310K  & \textcolor{blue}{23.89} & \textcolor{blue}{0.858} & \textcolor{blue}{0.104} \\
\midrule
\textbf{Ours}         & 624K  & \textcolor{red}{24.22} & \textcolor{red}{0.872} & \textcolor{red}{0.097} \\
\bottomrule
\end{tabular}
}
\caption{Quantitative comparison on the \textbf{LCDP} datasets for \textbf{exposure correction}.}
\vspace{-2mm}
\label{tab:table4}
\end{table}

Inspired by the recent work \cite{chen2022simple}, we utilize a three-layer U-net-style NAFNet as our backbone. Each layer in both the encoder and decoder contains two NAF-blocks. Instead of generating fixed size of bilateral grids, we set the grid size to 1/4 of the image resolution for 360p images in the PPR10K dataset, 1/32 for full-resolution images in the FiveK dataset, and 1/8 for all other images. This approach allows the model to balance efficiency and performance without sacrificing flexibility. Meanwhile, the grid's depth is fixed at 8.

For PPR10K datasets, 3DLUT-based methods employ a pre-trained ResNet-18 (11.7M) \cite{he2016deep} as their backbone, which cannot be directly integrated into our model. For a fair comparison, we increased our model's parameter count to a comparable level. Following the setting of \cite{liang2021ppr10k}, we leverage the portrait masks available in the dataset and assign a fivefold weight to the portrait regions when computing the MSE loss.

During training, we minimize the total loss end-to-end using the ADAM optimizer \cite{kingma2014adam}. The learning rate is updated using a cosine annealing schedule, with an additional decay factor of 0.1 to further enhance stability. The batch size is set to 1 across all datasets. For FiveK, PPR10K, and LCDP datasets, training consists of 225k, 500k, and 120k iterations, respectively. Our model is implemented based on the PyTorch framework. Specifically, the slicing operation and the application of the coefficients in the MLP are implemented using CUDA extension to speed up. All experiments are conducted on a NVIDIA RTX 4090 GPU.

\subsection{Comparison with State-of-the-Art Methods}
\noindent\textbf{Quantitative Comparisons.} We adopt PSNR and SSIM \cite{wang2004image} as the basic evaluation metrics, using the $L_2$-distance in CIE LAB color space($\Delta E$) for tone mapping and photo retouching tasks, and the LPIPS metric \cite{zhang2018unreasonable} for exposure correction tasks. We also include the human-centered PSNR ($PSNR^{HC}$)~\cite{liang2021ppr10k} and color difference ($\Delta E^{HC}$) on the PPR10K dataset. 
\begin{figure*}[t]
  \centering
  \setlength{\belowcaptionskip}{0pt}
  \includegraphics[width=\textwidth]{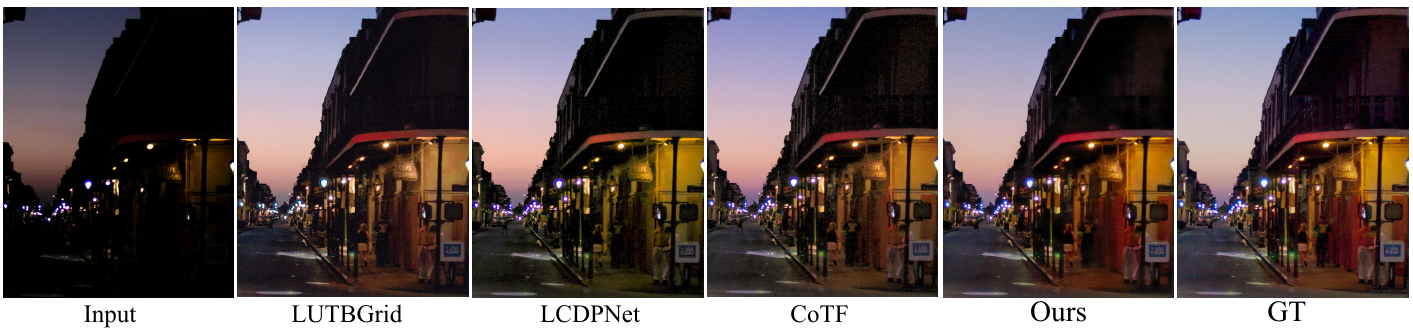}
  \caption{Visual comparison with state-of-the-art methods on the \textbf{LCDP} dataset for \textbf{exposure correction}.}
  \label{fig:figure6}
  \vspace{-1em}
\end{figure*}

Table \ref{tab:table1} shows comparison on FiveK dataset for tone mapping, Table \ref{tab:table2} and \ref{tab:table3} show comparisons on FiveK dataset and PPR10K dataset for photo retouching, Table \ref{tab:table4} is on LCDP dataset for exposure correction. It can be seen that our model exhibits the best performance on a range of tasks across three datasets. Speed comparison is also provided in Table \ref{fig:figure5}, we measure the inference time on 100 images and report the average.

On the FiveK dataset, our model achieves a significant lead (0.55dB) in terms of PSNR when handling full resolution images. It has also achieved state-of-the-art performance when directly applied to the exposure correction task. With our model outperforming in image quality metrics, it processes 4K images at a frame rate exceeding 30 FPS, meeting the requirements for real-time processing.

\begin{table}[htbp]
\centering
\label{tab:table5}
\resizebox{0.4\textwidth}{!}{%
\begin{tabular}{@{}l|cc|cc@{}}
\toprule
\multicolumn{1}{@{}l@{}}{\multirow{2}{*}{\textbf{Methods}}} & \multicolumn{2}{c}{\textbf{1920 $\times$ 1080}} & \multicolumn{2}{c}{\textbf{3840 $\times$ 2160}} \\
\cmidrule(lr){2-3}\cmidrule(lr){4-5}
\multicolumn{1}{@{}l@{}}{} & \multicolumn{1}{c}{\textbf{Time$\downarrow$}} & \multicolumn{1}{c}{\textbf{FPS$\uparrow$}} & \multicolumn{1}{c}{\textbf{Time$\downarrow$}} & \multicolumn{1}{c}{\textbf{FPS$\uparrow$}} \\
\midrule
HDRNet \cite{gharbi2017deep}    & 11.8  & 84.5  & 43.9  & 22.8  \\
3DLUT \cite{zeng2020learning}   & 0.89   & 1124 & 1.16   & 862 \\
CSRNet \cite{he2020conditional}  & 16.2  & 61.7  & 65.4  & 15.3  \\
LCDPNet \cite{wang2022local}     & 59.9  & 16.7  & 233 & 4.30   \\
LLF-LUT \cite{zhang2023lookup}   & /     & /    & 20.5  & 48.8  \\
LUTBGrid \cite{kim2024image}     & 1.66   & 602 & 3.14   & 319 \\  
CoTF \cite{li2024real}           & 8.02   & 125 & 28.3  & 35.4  \\
\midrule
\textbf{Ours}                   & 10.2  & 98.3  & 27.8  & 36.0  \\
\bottomrule
\end{tabular}
}
\caption{Running time (in millisecond) and frame rate (in FPS) comparison with state-of-the-art methods on high resolutions.}
\end{table}
\begin{table}[htbp]
\centering
\resizebox{0.45\textwidth}{!}{
  \begin{threeparttable}
    \begin{tabular}{ccccrr}
      \toprule
      Setting & Affine Trans & MLP & Grid Decomp & PSNR & SSIM \\
      \midrule
      1 & \checkmark &  &  & 25.53 & 0.935 \\
      2 &  & \checkmark &  & 25.70 & 0.939 \\
      3 & \checkmark &  & \checkmark & 25.63 & 0.937 \\
      4 &  & \checkmark & \checkmark & \textbf{25.83} & \textbf{0.941} \\
      \bottomrule
    \end{tabular}
  \end{threeparttable}
  }
\caption{Ablation study on two contributions of the BPAM.}
\label{tab:table6}
\end{table}
\noindent\textbf{Qualitative Comparisons.} From Figure \ref{fig:figure4} and \ref{fig:figure5}, it can be observed that our method consistently shows superior brightness and color performance. Particular in Figure~\ref{fig:figure5}, our approach enables clear discernment of the road and lawn in the darker rear areas, a level of detail that other methods struggle to achieve.
Figure \ref{fig:figure6} shows qualitative comparison for exposure correction task. In the illuminated areas, our method produces colors that are closest to the ground truth, whereas other methods exhibit varying degrees of color shift. More visual results can be found in the supplementary material.

\subsection{Ablation Studies}

We perform ablation studies for the tone mapping task on the FiveK dataset (480p) to verify the effectiveness of our proposed improvements. The results for each setting are listed in Table \ref{tab:table6}.  Setting 1 implements color mapping through the affine transformation conventionally employed in prior methods, yet exhibits relatively limited performance. Setting 2, which utilizes MLP for color transformation, demonstrates significant performance improvements over Setting 1, attributable to the MLP's enhanced non-linear modeling capabilities. Settings 3 and 4 further reveal that integrating the proposed grid decomposition strategy with both transformation approaches consistently yields additional performance gains. The grid decomposition strategy further exploits the potential of the bilateral grid, enabling more comprehensive utilization of color information and more precise estimation of grid parameters.
We provide more analysis in the supplementary material.

\subsection{Discussion}

Although our method delivers superior performance across various image enhancement tasks, its computational efficiency lags behind some state-of-the-art 3D LUT-based methods. This is a common challenge for spatial-aware methods. However, the high efficiency of 3D LUT-based methods comes at the cost of adaptability—they employ fixed-size 3D LUTs regardless of input resolution, which impairs their effectiveness on high-resolution images. Our approach achieves a considerably boosted performance on 4K images in the FiveK dataset, with a frame rate exceeding 30 FPS, striking a balance between computational efficiency and enhancement quality.

Moreover, we consider exposure correction to be a more challenging task due to the extreme illumination conditions it typically confronts. Direct application of 3D LUT-based method \cite{kim2024image} to this task yields suboptimal results. In contrast, our method shows consistent superiority in this task by implementing spatial-aware MLPs for nonlinear color mappings, addressing the intricate requirements of extreme exposure adjustments.

\section{Conclusion}
\label{sec:conclusion}

In this paper, we present a \textbf{B}ilateral Grid-based \textbf{P}ixel-\textbf{A}daptive \textbf{M}ulti-layer Perceptron (BPAM) framework for real-time image enhancement. The core idea is to leverage the bilateral grid, a data structure with spatial modeling capabilities, to assist in the estimation of MLPs parameters, using pixel-adaptive MLPs for color transformations. We further propose a grid decomposition strategy to alleviate the issue of insufficient utilization of color information when applying 3D bilateral grids to RGB images. Extensive experiments across three datasets demonstrate that our method outperforms previous methods and could meet the real-time efficiency requirements.

\noindent\textbf{Acknowledgment.} This work was supported by National Natural Science Foundation of China (No.62476051) and Sichuan Natural Science Foundation (No.2024NSFTD0041).

{
    \small
    \bibliographystyle{ieeenat_fullname}
    \bibliography{main}
}

\clearpage
\setcounter{page}{1}
\setcounter{section}{0}
\renewcommand{\thesection}{\Alph{section}}
\maketitlesupplementary
In Section \ref{A}, we present detailed information about our network architecture and the process of generating our bilateral grids. In Section \ref{B}, we describe the training configurations for each dataset. In Section \ref{C}, we provide further experimental analyses. In Section \ref{D}, we offer additional visual comparisons.

\section{Generation of the Bilateral Grids}
\label{A}

\begin{wrapfigure}{r}[0pt]{0pt} % "r" 表示图像靠右，宽度为0.5\linewidth
  \centering
  \includegraphics[width=0.35\linewidth]{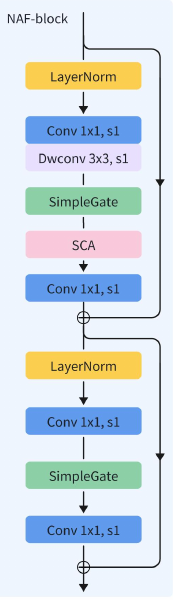}
  \caption{NAF-block}
  \label{fig:figure7}
\end{wrapfigure}
Our proposed method utilize a three-layer U-net-style NAFNet \cite{chen2022simple} as our backbone to generate the bilateral grids, with each layer in both the encoder and decoder comprising two NAF-blocks (see Figure \ref{fig:figure7}), and the number of blocks at the bottom level is the same. The width in NAF-block is set as 16. The feature map output from NAFNet maintains a channel numble equal to the model's width. We then apply the pixel unshuffle operation to reduce its resolution by a factor of 4 while expanding the channel count by 16 times. Next, two 1×1 convolutions are used to further adjust the channel dimensions, yielding two final feature maps. 

At this point, following HDRNet's \cite{gharbi2017deep} approach, we treat these feature maps as two bilateral grids whose third dimension has been unrolled
\begin{equation}
\mathcal{G}_{dc+z}[x,y]\leftrightarrow \mathcal{G}_c[x,y,z]
\end{equation}
where d represents the depth of the grid .We slice each map along the channel dimension into d parts and expand a new dimension, resulting in a three-dimensional bilateral grid in which each grid cell contains the required parameters.

\section{Experimental Details}
\label{B}
\subsection{FiveK}
Experiments on the FiveK dataset \cite{bychkovsky2011learning} are conducted on two different resolutions (480p and full resolution) and two tasks (retouching and tone mapping). For two format images (8-bit sRGB and 16-bit CIE XYZ) at 480p resolution, we adopt the dataset released by \cite{zeng2020learning}. For full-resolution images for tone mapping tasks, we utilize the dataset provided by \cite{zhang2023lookup}. 

The augmentations include random ratio cropping, random flipping and random rotating. For 480p and full-resolution images, we downsample by factors of 2 and 8, respectively, resulting in grids that are 1/8 and 1/32 of the original resolution. The training process consists of 225K iterations, we set the initial learning rate to 3e-4 and employ a cosine annealing schedule to gradually reduce it to 4e-6. 

\subsection{PPR10K}
Experiments on the PPR10K dataset \cite{liang2021ppr10k} are conducted on the 360p resolution for photo retouching task. The dataset also includes five augmented versions of the original training images, and the final training set comprises 53,250 images. For more details, please refer to \cite{liang2021ppr10k}.

The augmentations include random ratio cropping, random flipping. We do not perform downsampling on this dataset, as its native resolution is already low, resulting in grids that are 1/4 of the original resolution. 3D LUT-based approaches \cite{zeng2020learning,kim2024image,liang2021ppr10k} on this dataset employ ResNet-18 \cite{he2016deep}(11.7M) as their backbone, but this network cannot be used to generate our bilateral grids. For a fair comparison, we increased the depth of NAFNet to 4 layers, increased the width to 32 channels, and increased the number of bottom blocks to 4, thereby raising the model's parameter count to a comparable level (11.7M). The training process consists of 500K iterations, We set the initial learning rate to 2e-4 and employ a cosine annealing schedule to gradually reduce it to 2e-6. 

\subsection{LCDP}
Experiments on the LCDP dataset \cite{wang2022local} are conducted on the original resolution for exposure correction task. The augmentations include random ratio cropping, random flipping and random rotating. We downsample the images by a factor of 2, ultimately generating grids that are 1/8 of the original resolution. The training process consists of 120K iterations,, We set the initial learning rate to 4e-4 and employ a cosine annealing schedule to gradually reduce it to 2e-6. 

\section{Ablation Studies}
\label{C}
\noindent\textbf{Detailed Explanation about Grid Decomposition of Setting 3 in Table \ref{tab:table6}.} We divide the 12 coefficients of the
\begin{wrapfigure}[4]{r}[0pt]{0.6\linewidth}
    \vspace{-1.2em}
    \hspace*{-3.5em}
    \raggedleft
    \includegraphics[width=\linewidth,trim=0 5pt 0 5pt,clip]{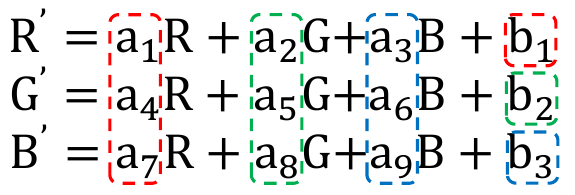}
    \vspace{-1em}
\end{wrapfigure}
  affine transformation into 3 parts and generate a 3-channel guidance map, where each channel takes its corresponding subset indicated by different colors.
  
\noindent\textbf{Selection of Guidance Maps}. We adopted the grid decomposition strategy to obtain multiple subgrids, which makes it necessary to use different guidance maps to steer the slicing operation for each corresponding subgrid. We designed two selection schemes and compare their performance for the tone mapping task on the FiveK dataset (480p), as shown in Table \ref{tab:table7}. In Setting 1, the grid decomposition strategy is not used, while in Settings 2 and 3 the grids are decomposed and different guidance map selection schemes are applied. The first scheme uses each input channel as its own guidance map to extract the corresponding weights, with a convolutional network fusing the input into a single channel for bias extraction. The second scheme, which is the one currently adopted, feeds the input into a convolutional network to generate a multi-channel guidance map, with each channel corresponding to a subgrid.
\begin{table}[htbp]
\centering
\caption{Comparison of two guidance maps selection schemes.}
\label{tab:table7}
\resizebox{0.45\textwidth}{!}{%
  \begin{threeparttable}
    \begin{tabular}{cccrr}
      \toprule
      Setting  & Scheme 1 & Scheme 2 & PSNR & SSIM \\
      \midrule
      1  &  &  & 25.70 & 0.939 \\
      2  & \checkmark &  & 25.78 & 0.940 \\
      3   &  & \checkmark & \textbf{25.83} & \textbf{0.941} \\
      \bottomrule
    \end{tabular}
  \end{threeparttable}
}
\end{table}

It can be seen that after applying grid decomposition, both guidance map selection schemes yield performance improvements, with the second scheme proving to be superior.
\begin{figure}[htbp]
  \flushright
  \includegraphics[width=1\linewidth]{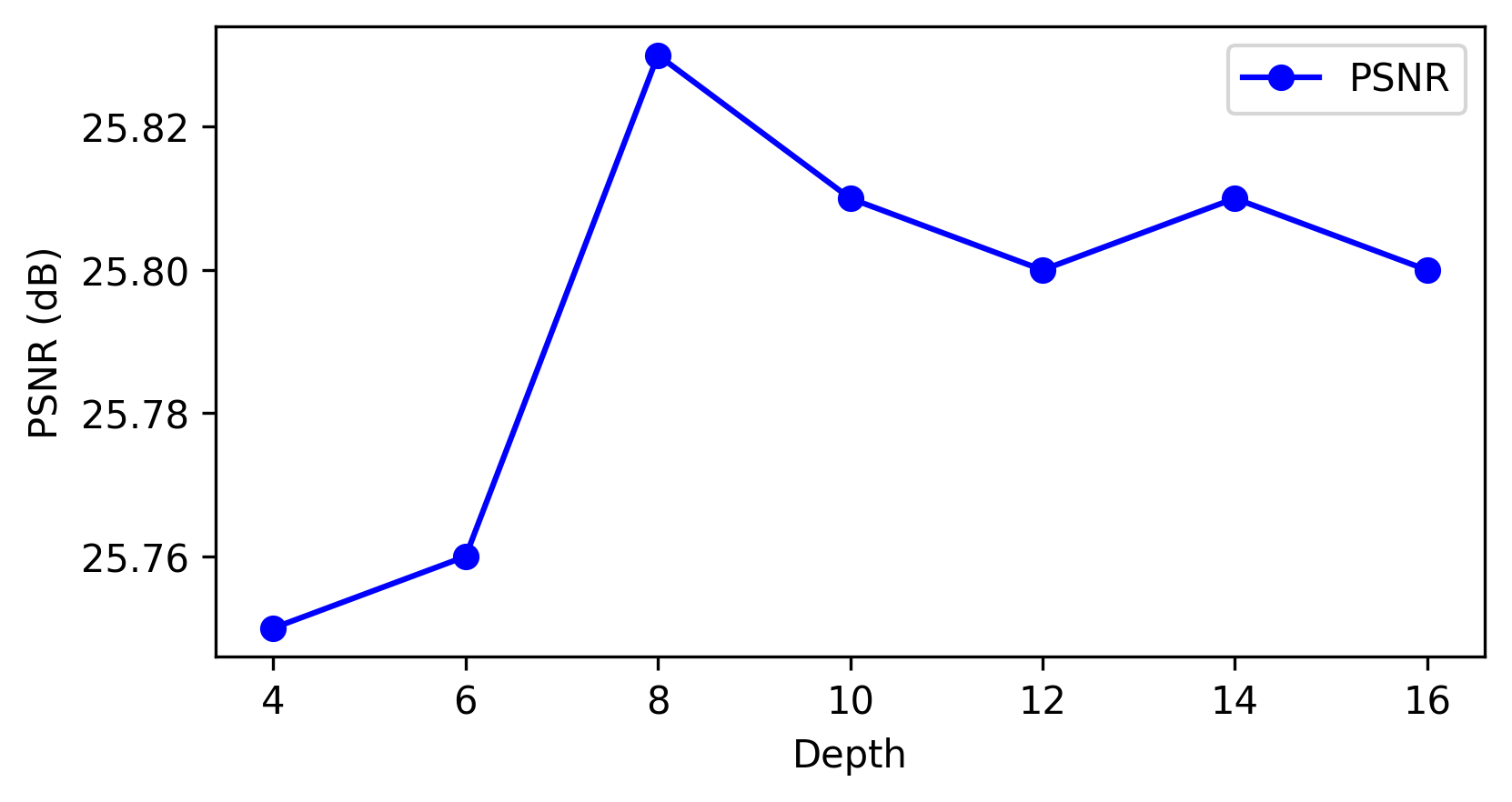}
  \caption{Ablation study on different numbers of bilateral grid depth.}
  \label{fig:figure8}
\end{figure}

\noindent\textbf{Depth of Bilateral Grid.} Depth in a bilateral grid controls the resolution along the intensity dimension. Generally speaking, higher depth captures finer details at the cost of efficiency, while lower depth improves speed but may lose subtle variations.
However, since the information in our bilateral grids is generated by the network, increasing the depth adds more parameters, which could make the model harder to train and doesn't necessarily improve performance. Therefore, We conduct experiments for tone mapping on the FiveK dataset to determine the optimal settings, results can be seen in Figure \ref{fig:figure8}. It can be observed that the model achieves its best performance when the grid depth is 8, so we ultimately adopt this setting.

\noindent\textbf{Ablation study about MLP depth and channels.} We train models with increased channel numbers and depth of MLP. According to the table \ref{tab:table8}, increasing the number of intermediate layers in the MLP degrades model performance. This occurs because the dramatic increase in internal parameters makes it difficult for the model to learn effectively. While increasing the depth of the MLP also contributes to this issue, it enhances the model's nonlinear capabilities, leading to some improvement in performance.
Further increasing the complexity of MLPs is not able to significantly improve mapping ability but leads to more computational costs. 
\begin{table}[htbp]
\centering
\caption{Ablation study about MLP depth and channels.}
\label{tab:table8}
\resizebox{0.35\textwidth}{!}{%
  \begin{tabular}{c|c|c|r}      % 保留竖线格式
    \toprule[1.2pt]
    Setting  & Params & Time(4K)  & PSNR \\
    \midrule[0.8pt]
    3-8-3    & 624K   & 27.8 ms   & 25.83 \\
    3-16-3   & 741K   & 42.8 ms   & 25.81 \\
    3-8-8-3  & 779K   & 48.3 ms   & 25.87 \\
    \bottomrule[1.2pt]
  \end{tabular}
}
\end{table}

\section{More Qualitative Results}
\label{D}
\begin{figure*}[t]
  \centering
  \setlength{\belowcaptionskip}{0pt}
  \includegraphics[width=\textwidth]{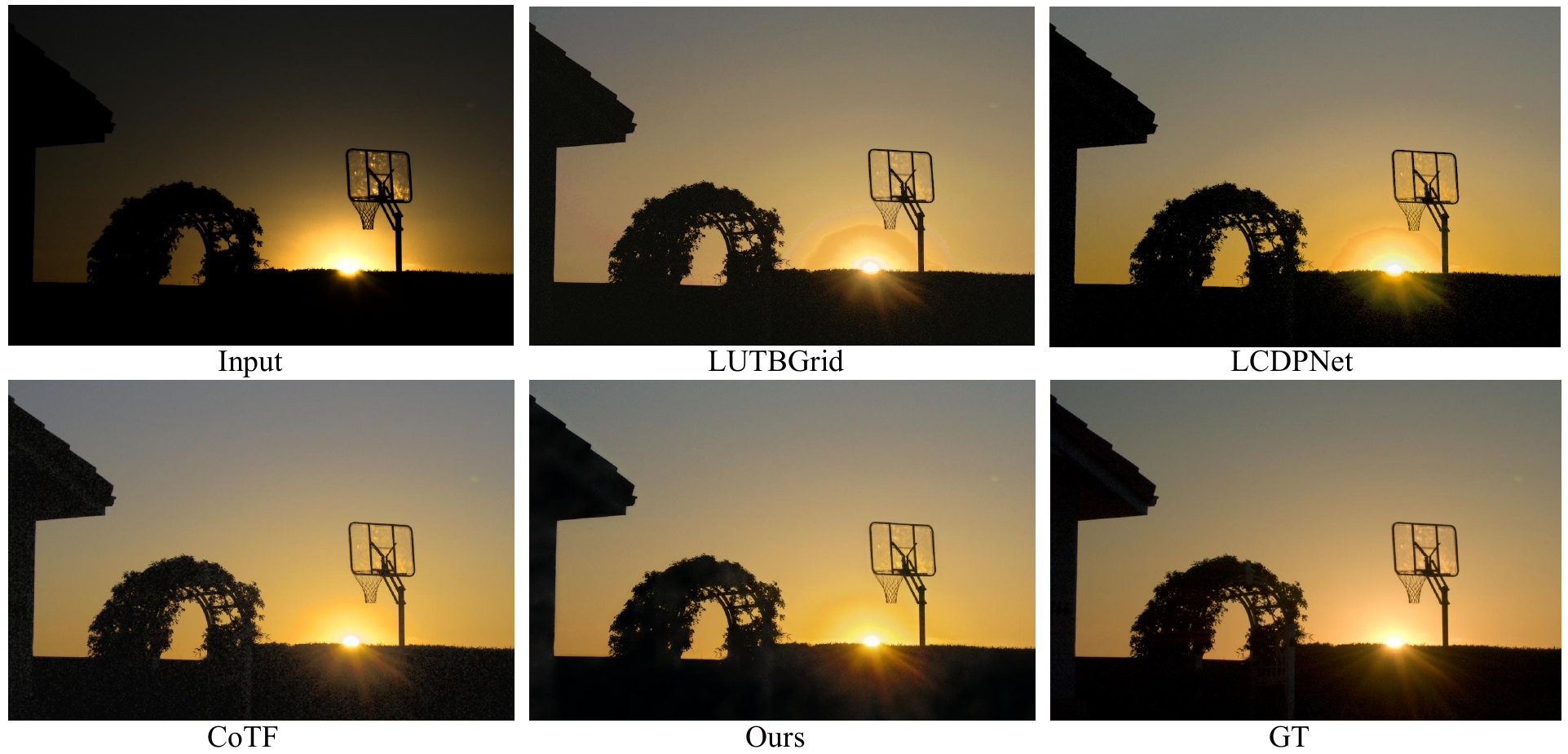}
  \caption{Visual comparison with state-of-the-art methods on the LCDP dataset for exposure correction.}
  \label{fig:figure9}
\end{figure*}

\begin{figure*}[t]
  \centering
  \setlength{\belowcaptionskip}{0pt}
  \includegraphics[width=\textwidth]{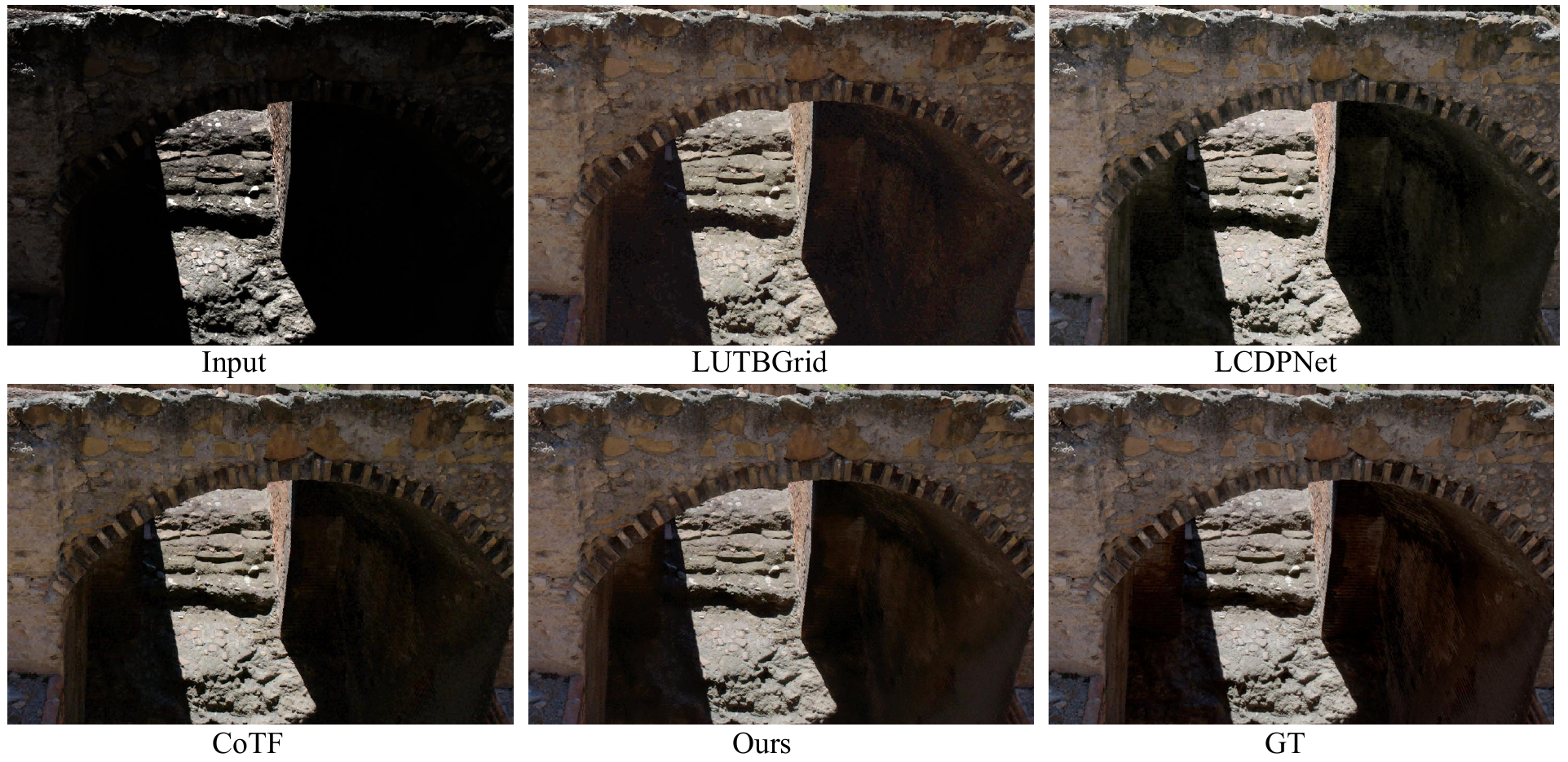}
  \caption{Visual comparison with state-of-the-art methods on the LCDP dataset for exposure correction.}
  \label{fig:figure10}
\end{figure*}

\begin{figure*}[t]
  \centering
  \setlength{\belowcaptionskip}{0pt}
  \includegraphics[width=\textwidth]{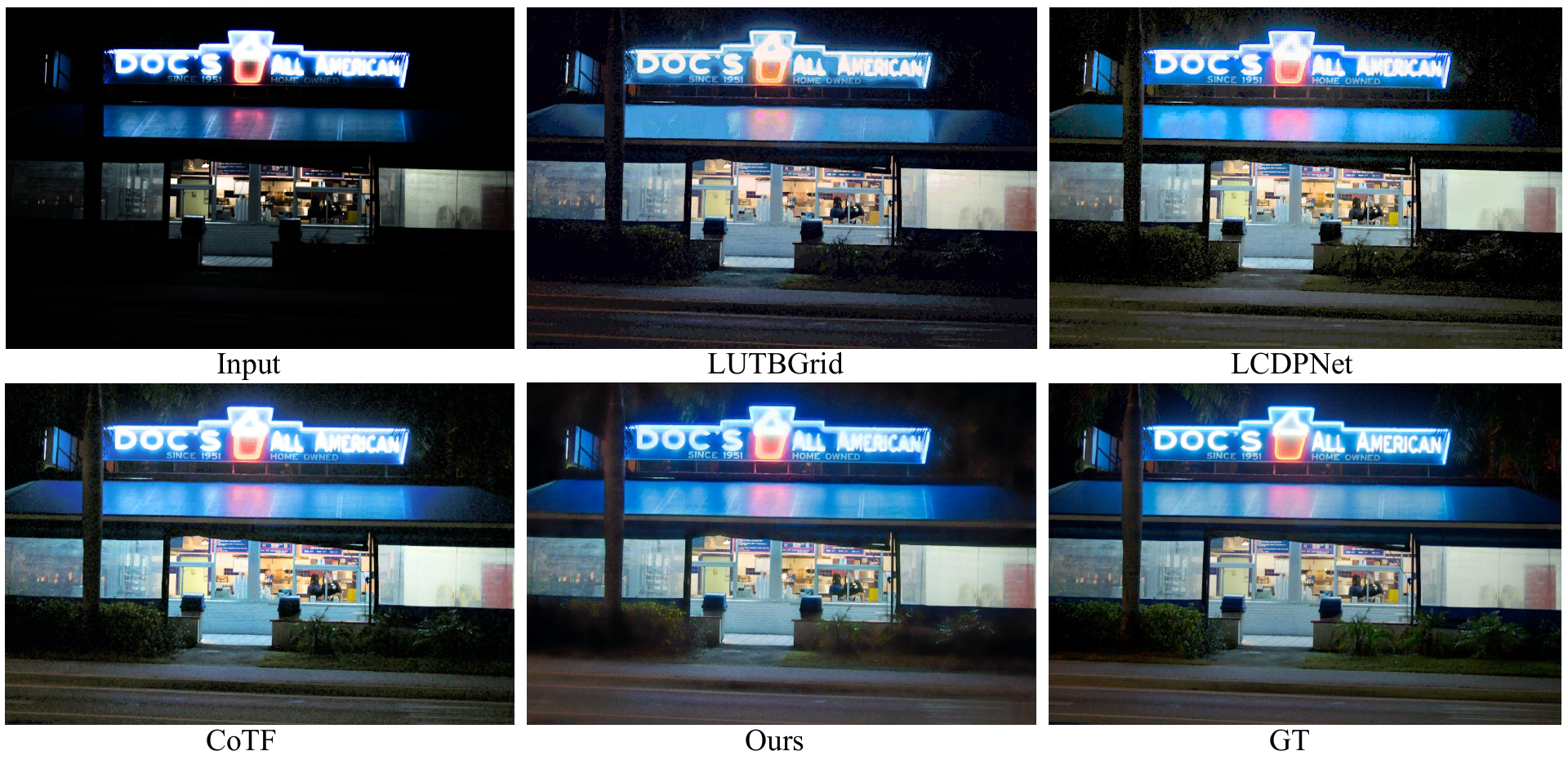}
  \caption{Visual comparison with state-of-the-art methods on the LCDP dataset for exposure correction.}
  \label{fig:figure11}
\end{figure*}

\begin{figure*}[t]
  \centering
  \setlength{\belowcaptionskip}{0pt}
  \includegraphics[width=\textwidth]{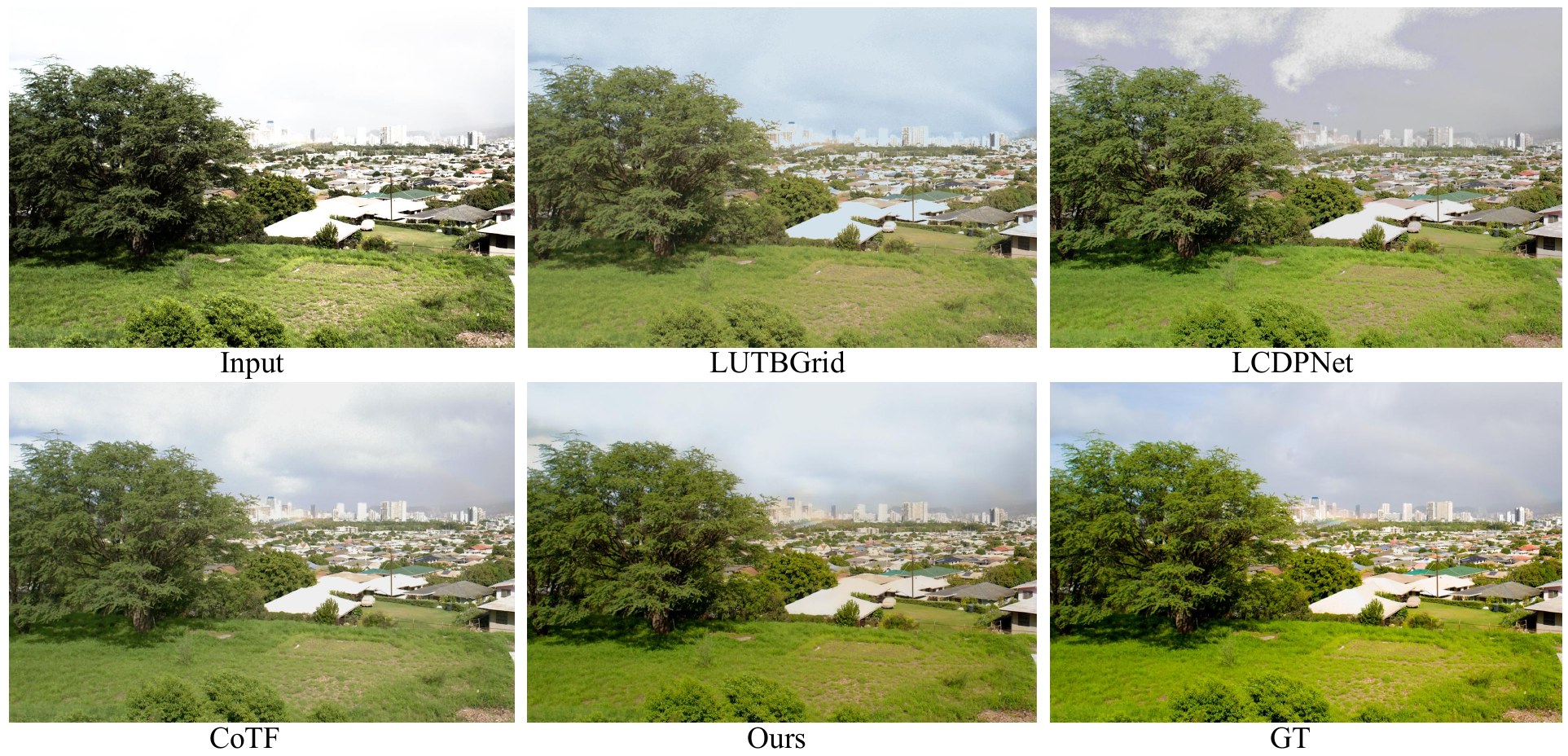}
  \caption{Visual comparison with state-of-the-art methods on the LCDP dataset for exposure correction.}
  \label{fig:figure12}
\end{figure*}

\begin{figure*}[t]
  \centering
  \setlength{\belowcaptionskip}{0pt}
  \includegraphics[width=\textwidth]{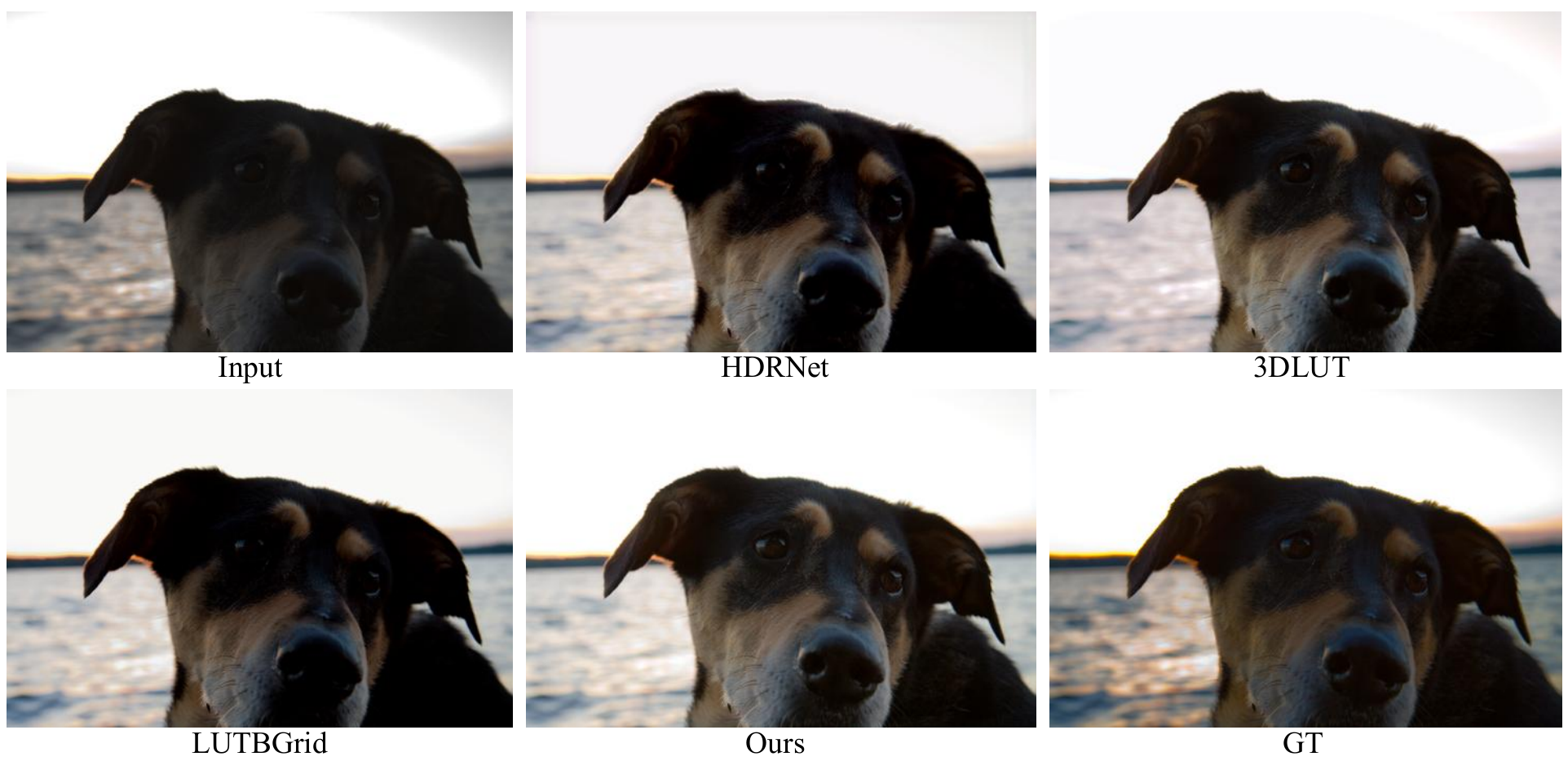}
  \caption{Visual comparison with state-of-the-art methods on the FiveK dataset (4K) for tone mapping.}
  \label{fig:figure13}
\end{figure*}

\begin{figure*}[t]
  \centering
  \setlength{\belowcaptionskip}{0pt}
  \includegraphics[width=\textwidth]{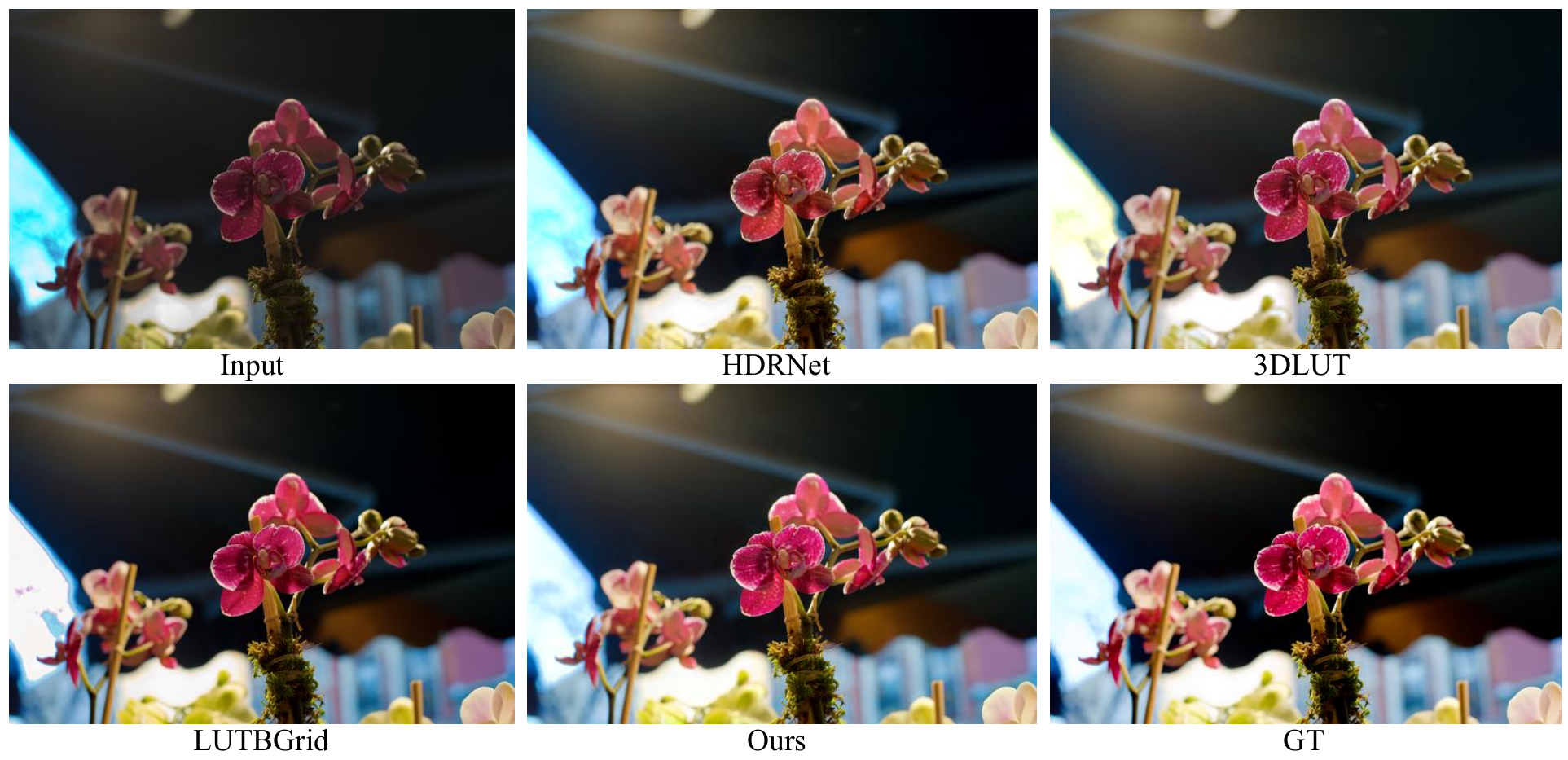}
  \caption{Visual comparison with state-of-the-art methods on the FiveK dataset (4K) for tone mapping.}
  \label{fig:figure14}
\end{figure*}

\begin{figure*}[t]
  \centering
  \setlength{\belowcaptionskip}{0pt}
  \includegraphics[width=\textwidth]{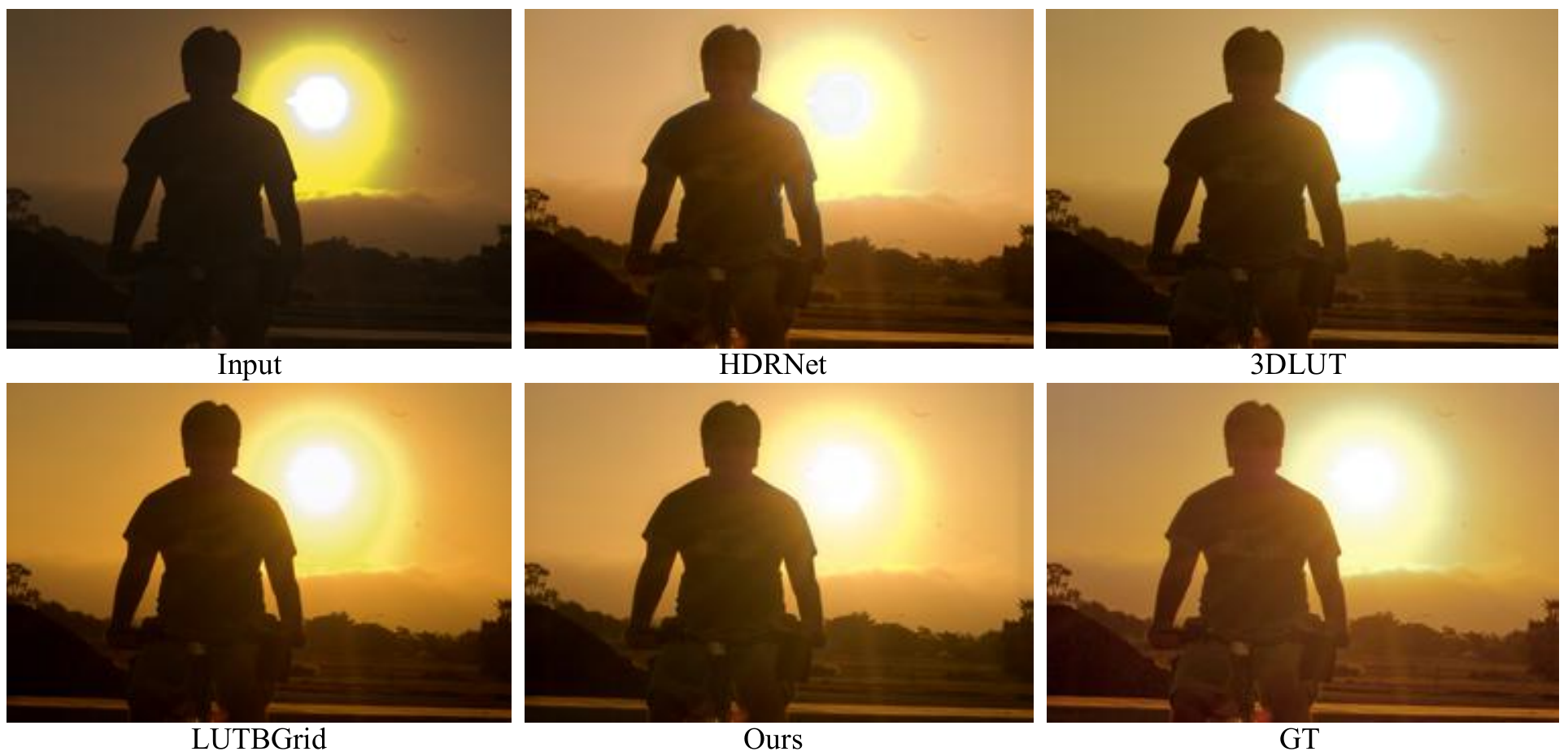}
  \caption{Visual comparison with state-of-the-art methods on the FiveK dataset (4K) for tone mapping.}
  \label{fig:figure15}
\end{figure*}

\begin{figure*}[t]
  \centering
  \setlength{\belowcaptionskip}{0pt}
  \includegraphics[width=\textwidth]{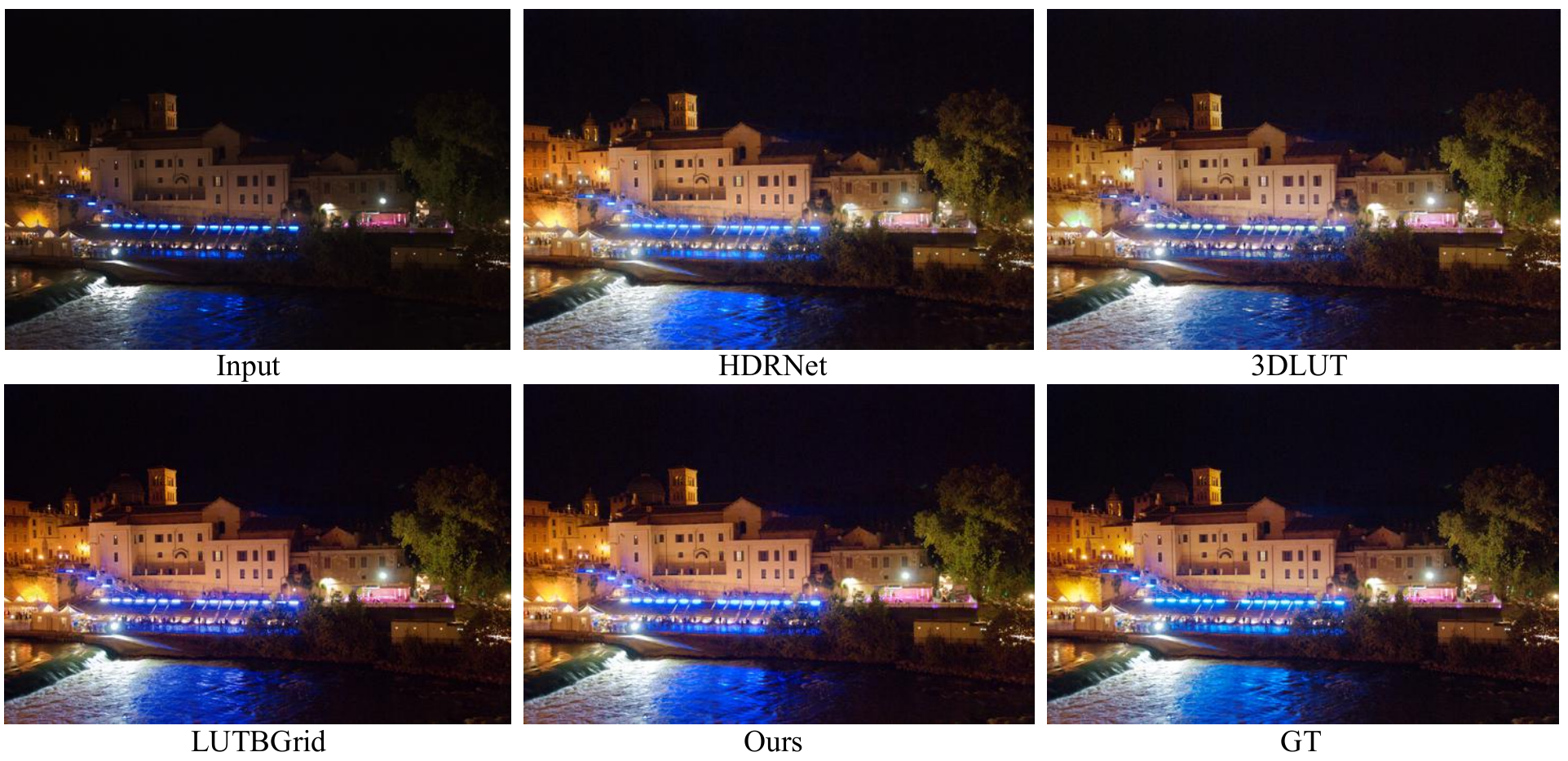}
  \caption{Visual comparison with state-of-the-art methods on the FiveK dataset (4K) for tone mapping.}
  \label{fig:figure16}
\end{figure*}

We provide additional visual comparisons on the LCDP dataset in Figure \ref{fig:figure9}, \ref{fig:figure10}, \ref{fig:figure11}, \ref{fig:figure12} and on the FiveK (4K) dataset in Figure \ref{fig:figure13}, \ref{fig:figure14}, \ref{fig:figure15}, \ref{fig:figure16}.

\end{document}